\documentclass[]{spie}  

\usepackage{amsmath,amsfonts,amssymb}
\usepackage{graphicx}
\usepackage{adjustbox}
\usepackage[colorlinks=true, allcolors=blue]{hyperref}

\title{Longitudinal Variability Analysis on Low-dose Abdominal CT with Deep Learning-based Segmentation}

\author[a]{Xin Yu}
\author[b]{Yucheng Tang}
\author[a]{Qi Yang}
\author[a]{Ho Hin Lee}
\author[a]{Riqiang Gao}
\author[a]{Shunxing Bao}
\author[c]{Ann Zenobia Moore}
\author[c]{Luigi Ferrucci}
\author[a,b]{Bennett A. Landman}
\affil[a]{Computer Science, Vanderbilt University, Nashville, TN, USA}
\affil[b]{Electrical and Computer Engineering,  Vanderbilt University, Nashville, TN, USA}
\affil[c]{National Institute on Aging, Baltimore, MD}

\begin{document} 
\maketitle

\begin{abstract}
Metabolic health is increasingly implicated as a risk factor across conditions from cardiology to neurology, and efficiency assessment of body composition is critical to quantitatively characterizing these relationships.  2D low dose single slice computed tomography (CT) provides a high resolution, quantitative tissue map, albeit with a limited field of view. Although numerous potential analyses have been proposed in quantifying image context, there has been no comprehensive study for low-dose single slice CT longitudinal variability with automated segmentation. We studied a total of 1816 slices from 1469 subjects of Baltimore Longitudinal Study on Aging (BLSA) abdominal dataset using supervised deep learning-based segmentation and unsupervised clustering method. 300 out of 1469 subjects that have two year gap in their first two scans were pick out to evaluate longitudinal variability with measurements including intraclass correlation coefficient (ICC) and coefficient of variation (CV) in terms of tissues/organs size and mean intensity. We showed that our segmentation methods are stable in longitudinal settings with Dice ranged from 0.821 to 0.962 for thirteen target abdominal tissues structures. We observed high variability in most organ with ICC$\textless$0.5, low variability in the area of muscle, abdominal wall, fat and body mask with average ICC$\geq$0.8. We found that the variability in organ is highly related to the cross-sectional position of the 2D slice. Our efforts pave quantitative exploration and quality control to reduce uncertainties in longitudinal analysis.
\end{abstract}


\section{INTRODUCTION}
\label{sec:intro}  

Body composition is associated with cardiovascular disease, diabetes and cancers \cite{oh2021changes}. Body composition measurements help monitor the changes related to growth and disease progressions \cite{kuriyan2018body}. Computed tomography is widely used to assess body composition \cite{de2011body}. Single abdominal slice is commonly selected to measure body distribution instead of CT volume to reduce radiation absorbed by human body \cite{zopfs2020single}.

Much effort has been made to analyze the longitudinal relationship between changes of body composition and progression of disease or habitus. Raguso \textit{et al.} \cite{raguso20063} conducted a 3-year longitudinal study on body composition to illustrate the role of between physical exercise and body composition. Jackson \textit{et al.} \cite{jackson2012longitudinal} studied longitudinal changes in body composition associated with aging with healthy men aged 20-96 years old. Hughes \textit{et al.} \cite{hughes2002longitudinal} perfomed longitudinal study to examine the body composition change in elderly people. Rossi \textit{et al.} \cite{rossi2008body} conducted 7-year follow-up evaluation for the relationship between body composition change and pulmonary function. However, no comprehensive study has been done for longitudinal variability analysis using body compositions derived from low-dose single slice CT images.


Precise segmentation of soft tissues and organs within 2D abdominal slices is the prerequisite to perform longitudinal variability analysis. Deep learning-based method show its efficacy on semantic segmentation task. Chen \textit{et al.} \cite{chen2017rethinking} proposed Deeplab-v3 to exploit broader field of views. Ronneberger \textit{et al.} \cite{ronneberger2015u} demonstrated the effectivity on skip connection with U shape architecture in preserving edge-wise feature on biomedical segmentation.
Inspired by these works, we designed our pipeline to segment a total of thirteen tissues including organs, muscle, fat, abdominal wall and body mask on 2D axial abdominal slices. We designed deep learning-based segmentation methods for multi-organ, muscle and abdominal wall segmentation and unsupervised threshold/clustering based methods for body mask and fat segmentation. The methods were trained and evaluated on 1469 subjects.  300 subjects that have 2 year gap on their first two scans were performed statistical analysis including intraclass coefficient (ICC) and coefficient of variation (CV) were performed on their segmentation masks to analyze the longitudinal variability. Our contributions can be summarized as below:    
\begin{enumerate}
\item We designed a segmentation pipeline to assess a total of 13 tissues and organs using 2D low-dose single slice CT images,
\item To the best of our knowledge, this is the first work of performing longitudinal variability analysis using low-dose abdominal single slice CT images.
\end{enumerate}

\section{Methods}

\begin{figure*}[h!]
  \centering
  \includegraphics[width=\textwidth]{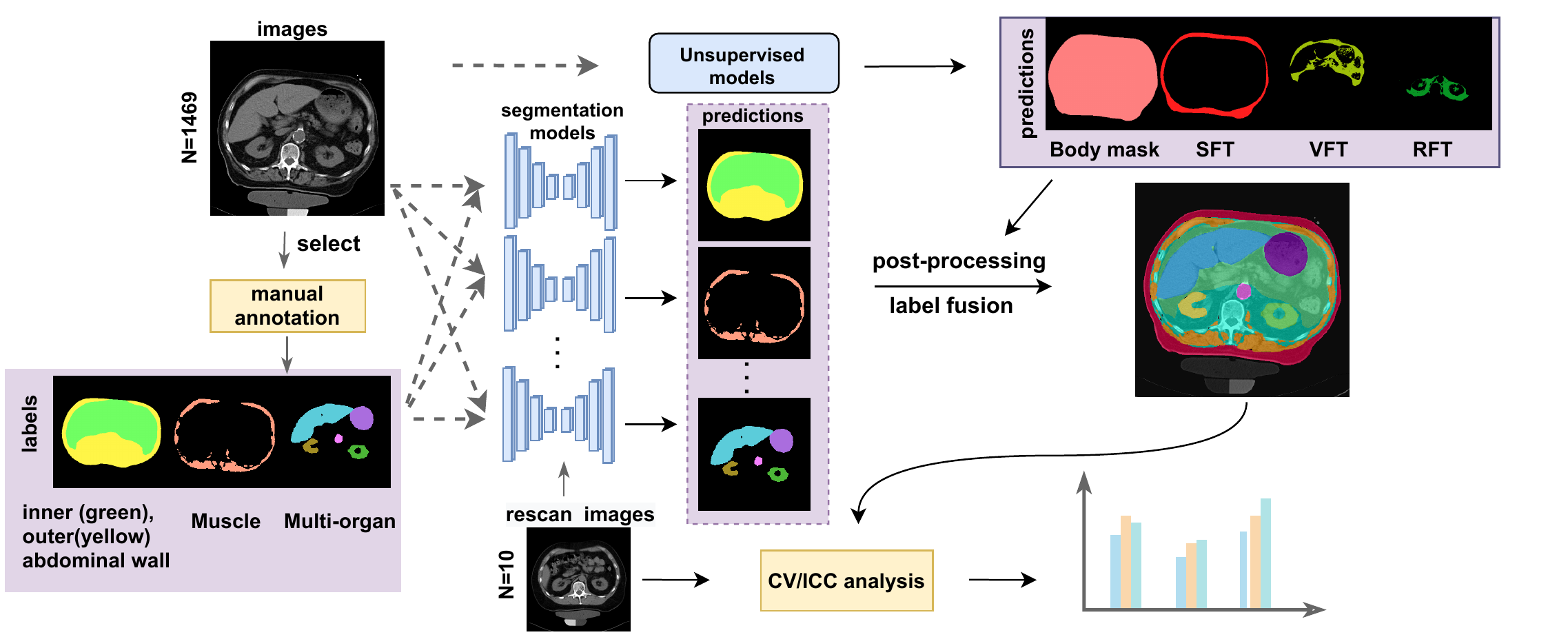}
\caption{The pipeline of the proposed framework. The dataset contains 1816 2D single CT slices from 1469 subjects. 1314, 300 and 1175 subjects were selected to perform manual annotations to provide the ground truth labels for the inner/outer abdominal wall, muscle, and multi-organ, respectively. The images and ground truth labels were fed into different segmentation models to estimate the corresponding prediction. The images were also fed into unsupervised clustering models for the body mask and fat segmentation.Segmentation results were used to perform CV and ICC analysis. }
\label{fig:fig1}
\end{figure*}

As shown in Figure~\ref{fig:fig1}, our longitudinal analysis consists of two sections: 1) abdominal organ and tissue segmentation and 2) longitudinal variability analysis. 
\subsection{Low-dose Single Slice Segmentation}
We performed multi-organ, abdominal wall and muscle segmentation on single abdominal slice. Deeplab-v3 \cite{chen2017rethinking} with ResNet101 \cite{he2016deep} were adopted as the backbone for the multi-organ and abdominal wall segmentation. It is challenging to perform automatic segmentation on 2D single slice CT due to a lack of 3D medical context. Here, we only included large organs (spleen, right/left kidney, liver, stomach and aorta) in this study. For the multi-organ segmentation, the total subjects were divided into 80\% and 20\% for training/validation and test, respectively. For inner/outer abdominal wall segmentation, we focused on segmentation of the retroperitoneal area. The retroperitoneal area is an anatomical area which is located posterior to the abdomen cavity including the left and right kidney and aorta. A small population of the annotated dataset as shown in Figure~\ref{fig:fig1} was used to train an initial segmentation network. Cases that segmentation model failed on the retroperitoneal area were picked out. A small portion of these cases were corrected manually and added to the training dataset. This process was iterated until all the pseudo labels for the unlabeled data can segment the retroperitoneum area correctly. This iterative process actively learned to adapt the outlier cases training with the refined pseudo labels \cite{yu2022accelerating}.
A U-Net \cite{ronneberger2015u} was adopted for muscle segmentation. The U-Net was designed to have 5 down-sampling blocks and 5 up-samplings blocks with a starting filter of 8. The objective function of supervised segmentation training is:
\begin{equation}
L= -\sum^n_{i=1}t_ilog(p_i),
\end{equation}
where $t_i$ is the ground truth label and $p_i$ is the probability for the $i^{th}$ class.
\subsection{Unsupervised Fat Tissue Segmentation}
We presented unsupervised methods for body tissue extraction. The body mask and fat tissue were obtained by using a two-stage fuzzy-c means \cite{bezdek1984fcm}. The fuzzy clusters classify pixels into two categories based on the CT original intensities by minimizing the loss function as below: 

\begin{equation}
m_\text{k}(x)= (\sum^C_{j=1} \frac{||{\mu(x)-\mu_k}||^2}{||{\mu(x)}-\mu_j||})^{-1},
\end{equation}

\begin{equation}
L = \sum^C_{k=1}\sum_x m_\text{k}(x)||\mu(x)-\frac{\sum_x m_k(x)\mu(x)}{\sum_x \mu(x)}||,
\end{equation}
where $x$ is the pixel coordinates in the image, $C$ is the number of clusters, $m_k(x)$ is the membership of coordinate $x$ in class $k$ and $\mu_k$ is the centroid of class $k$,  $\mu(x)$ is the intensity value at position $x$. 

After obtaining body mask by thresholding, we detected the fat tissue in following two iterations. First, we computed the initial membership value and cluster centroids. Second iteration identified the brighter and darker pixels by membership thresholds. Then we clustered the fat area. As in Figure~\ref{fig:fig1}, subcutaneous fat tissue (SFT) was the complementary set of the body mask and abdominal wall. Visceral fat tissue (VFT) and retroperitoneum fat tissue (RFT) were within the inner and outer abdominal wall respectively. 

\subsection{Longitudinal Variability Measurements}
The results of multi-organs, abdominal wall, muscle, body mask and three different fat segmentation were first refined by removing the tiny connected components with fewer than 25 pixels. The removed pixels were filled with the nearest label. 
The final prediction was fused with the order of multi-organ, abdominal wall and body mask on overlapping labels.
Area in square millimeters of each tissues was calculated by multiplying the amount of pixel and the pixel spacing. Mean intensity represented the mean HU value inside the segmentation mask. We used ICC and CV to evaluate longitudinal variability and served as a reference to measure whether the tissue can be used as a metric. ICC and CV expressed the similarity level across the observations of scans and the differences between scans, respectively \cite{wittensinter}. Our study used ICC and CV to evaluate the difference of mean intensity and area of each same segmented tissue from different scans. 
ICC was computed using a two-way mixed method:
\begin{equation}
ICC = \frac{\sigma^2_A}{\sigma^2_A + \sigma^2_w},
\end{equation}
where $\sigma^2_A$ is the variance amongst groups and $\sigma^2_w$ is the variance within groups \cite{wittensinter}. ICC range from 0 (no agreement) to 1 (absolute agreement).

The CV is defined as the ratio of the standard deviation to the mean of measurements:
\begin{equation}
CV (\%) = \frac{\sigma}{\mu} \times 100,
\end{equation}
since HU ranges [-1024, 3071], to avoid negative values when calculating intensity, we added 1024 to each value.

\begin{table*}[ht]
\caption{The results of segmentation for the supervised method in terms of Dice. ICC and CV for both area and intensity. The intensity was added 1024 when calculating CV intensity to avoid negative values. }
\begin{adjustbox}{width=\textwidth}
\begin{tabular}{lccccccccccccc}
\hline
Organ             & Spleen & R. kidney & L. kidney & Liver  & Stomach & Aorta & Muscle & Inner wall & Outer wall & SFT              & VFT              & RFT              & Body mask        \\ \hline\hline
Dice              & 0.821  & 0.850     & 0.845     & 0.900  & 0.856   & 0.822 & 0.936  & 0.962      & 0.958      & \textbackslash{} & \textbackslash{} & \textbackslash{} & \textbackslash{} \\
ICC area          & 0.186  & 0.657     & 0.478     & 0.662  & 0.003   & 0.461 & 0.920  & 0.805      & 0.925      & 0.938            & 0.854            & 0.887            & 0.968            \\
ICC intensity     & 0.202  & 0.378     & 0.372     & 0.327  & 0.297   & 0.578 & 0.924  & 0.545      & 0.856      & 0.805            & 0.600            & 0.694            & 0.754            \\
CV area (\%)      & 20.786 & 10.558    & 11.670    & 22.244 & 29.091  & 8.851 & 3.968  & 4.964      & 3.408      & 6.306            & 11.092           & 7.188           & 2.084            \\
CV intensity (\%) & 0.683  & 0.177     & 0.199     & 0.317  & 4.241   & 0.301 & 0.128  & 1.801      & 0.333      & 0.387            & 0.448            & 0.259            & 0.703            \\ \hline
\end{tabular}
\end{adjustbox}
\label{table:table1}
\end{table*}
\section{experiments and results}
\label{sec:typestyle}
\subsection{Dataset and Implementation Details}
A total of 1816 2D single slice abdominal CT images from 1469 subjects of the Baltimore Longitudinal Study on Aging (BLSA) cohort were used in this work. The data had been de-identified and approved by the Institutional Review Board (IRB). 1175 (80\%) subjects were used for multi-organ segmentation training/validation and 294 (20\%) for testing. For the muscle segmentation, 300 subjects were used for training/validation and 60 subjects were used for test. When conducting the abdominal segmentation, 100 subjects were left out as the external test set. The training was initially started with 100 annotated subjects. The process of training and refining the outlier pseudo labels was iterated to ensure the model's capability of distinguishing the retroperitoneum area. The data had the size of 512 $\times$ 512 with a pixel spacing of 0.9766 mm $\times$ 0.9766 mm. The data were processed with the soft tissue window in the range of [-125, 275] HU as in \cite{zhou2019prior} and rescaled to [0, 255]. During online data augmentation, the images were randomly flipped with 0.5 probability and resized with range [0.5, 2.0]. Segmentation results were evaluated in terms of the Dice coefficient 
(Table ~\ref{table:table1}).

\subsection{Longitudinal Analysis}
300 out of 1469 subjects whose first two scans have the time interval of two years were involved in the analysis. Within 300 subjects, 90, 72 and 197 subjects were not involved into the training of multi-organ, abdominal wall and muscle segmentation, respectively, and thus only these subjects were included in its corresponding analysis. Segmentation models were applied on both two scans. Area and mean intensity for each tissue/organ were calculated based on the segmentation masks and they were analyzed using ICC and CV. The results were shown in Table ~\ref{table:table1}.
Muscle showed an average ICC $\geq$ 0.90 for both area and intensity. The area of the abdominal wall, body mask and fat also show high ICC $\textgreater$ 0.80.
\begin{figure}[htb]
  \centering
  \includegraphics[width=\textwidth]{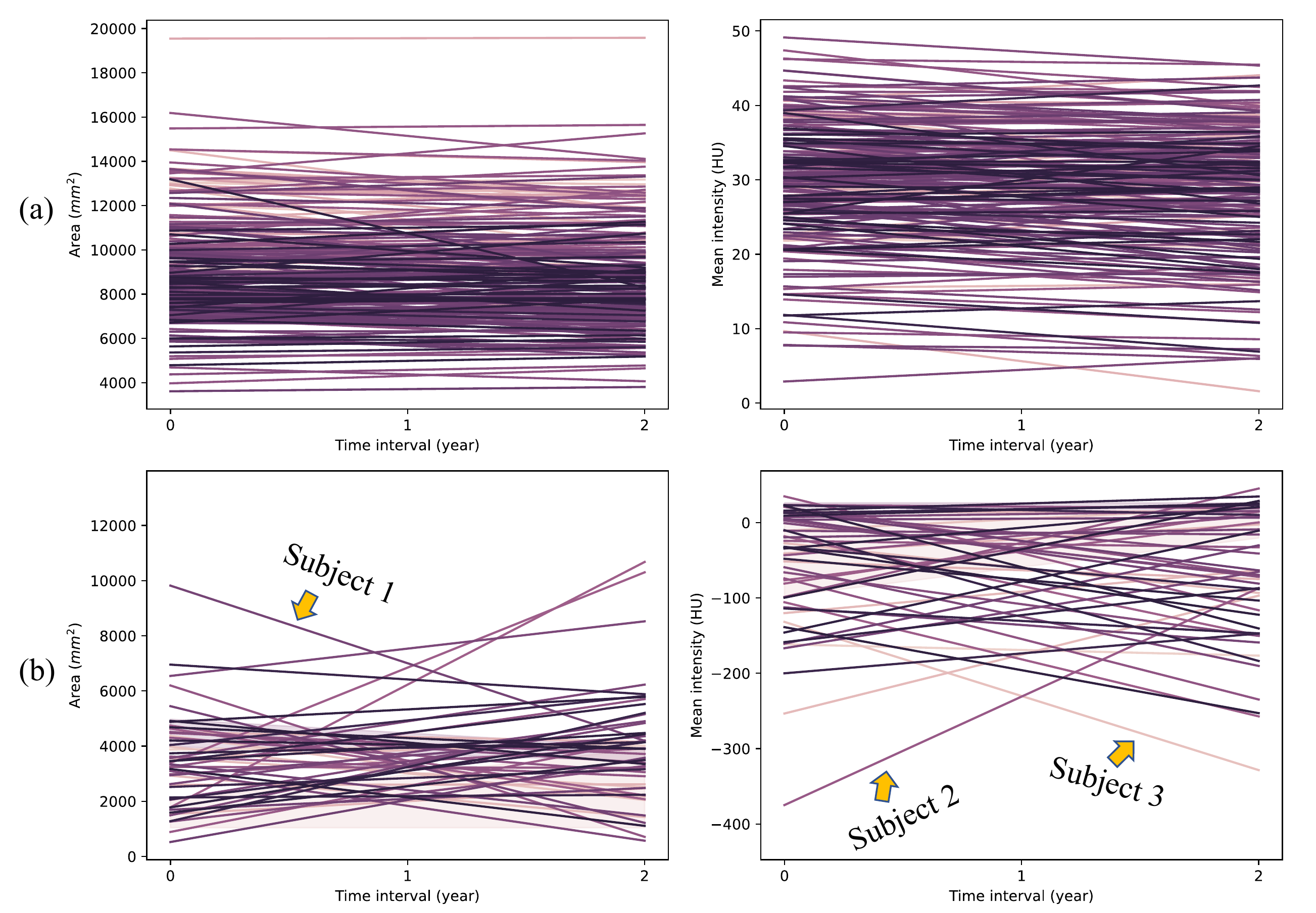}
\caption{The longitudinal spaghetti plot of (a) muscle and (b) stomach of area and mean intensity. 197 subjects involved (a) and 90 subjected involved in (b). (a) shows low longitudinal variability in both area and intensity. (b) shows high longitudinal variability in both area and intensity. Orange arrows represent outliers. }\label{fig:fig2}
\end{figure}

All the organs had ICC $\textless$ 0.70. This indicated that organs had low agreement. This low agreement related to the different vertebral level 2D axial slice being scanned (Figure~\ref{fig:fig3}). The CV served as a complementary measurement showed the same trend as the ICC. The organ/tissue with high ICC had a low CV, vice versa. Figure~\ref{fig:fig2} shows the longitudinal spaghetti plot between stomach and muscle of the two scans for 90 and 197 subjects, respectively. The area/intensity of the muscle showed a low longitudinal variability in two years while high longitudinal variability was found in stomach. The segmentation results for the outliers in Figure~\ref{fig:fig2} were showed in Figure~\ref{fig:fig3}.

\section{DISCUSSION AND CONCLUSION}
Dice score in Table~\ref{table:table1} shows our methods are stable in longitudinal single slice segmentation. High ICC of area of muscle, abdominal wall, fat and body mask indicates low longitudinal variability within 2 years. Muscle and fat are essential parts of body composition, the high agreement in muscle and fat may further indicate the feasibility of using low-dose single slice CT images to analyze the body composition. 

\begin{figure}[htb]
\begin{minipage}[b]{1.0\linewidth}
  \centering
  \includegraphics[width=\textwidth]{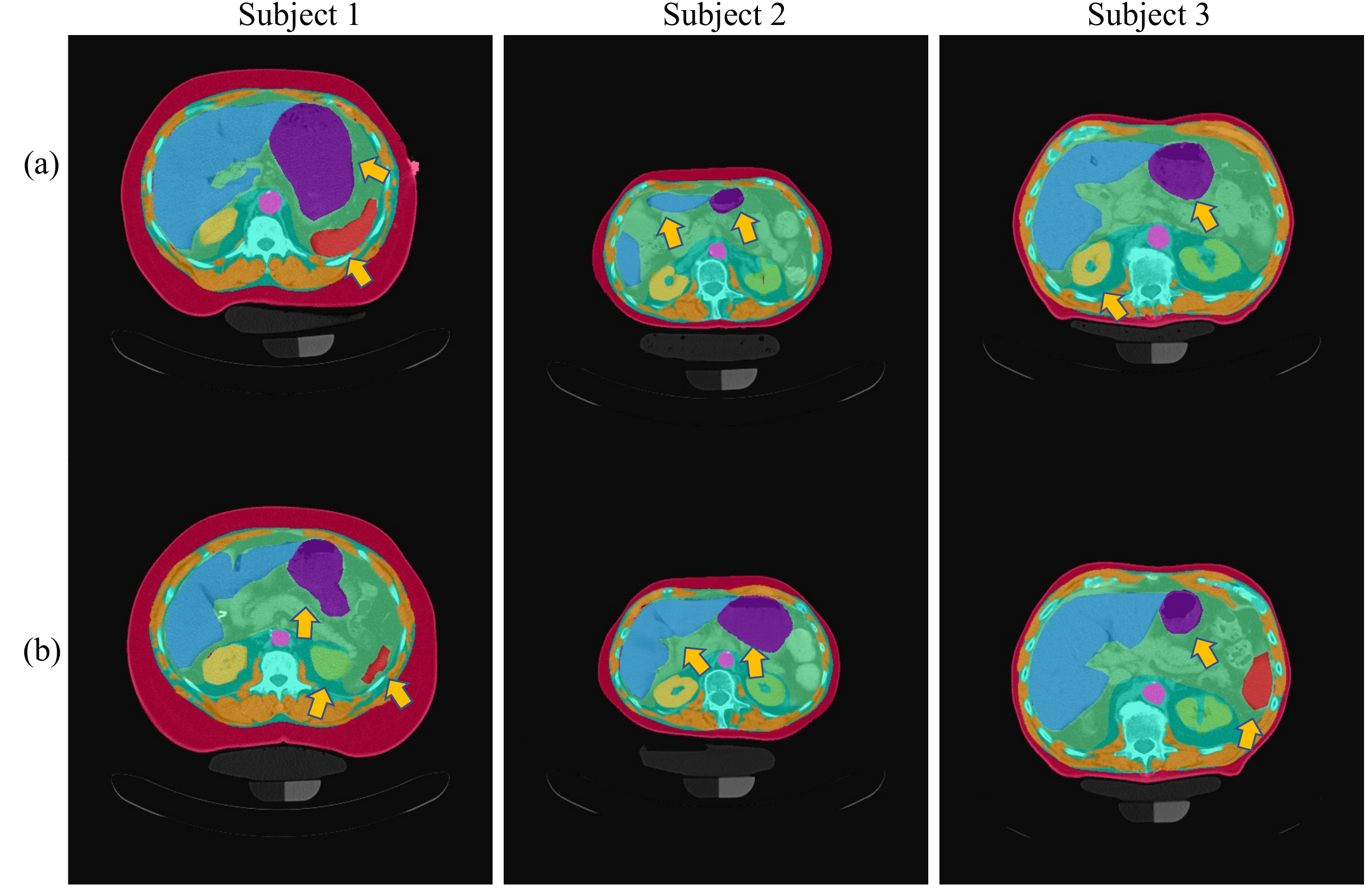}
\caption{Visualization of the fused segmentation results of the outliers in Figure~\ref{fig:fig2}. (a) shows the images and corresponding segmentation in the first scan and (b) shows the those after two years. The orange arrows indicate variations of 2 scans within subject.}\label{fig:fig3}
\end{minipage}
\end{figure}

However, the ICC of area of most of the organs is low, and this may be related to the abdominal position of the captured slice. From Figure~\ref{fig:fig3}, the organs of the same subject have different morphology and size across two scans. The size difference is introduced by the variation of the slice cross-sectional position instead of the time gap. However, the organs with low ICC and high CV do not necessarily imply a low agreement and high variation within 2 years.

The ICC of intensity is low for most of the structure/tissue could because that the HU unit is standardised for each tissue. The inter-subject variation for the same tissue/structure might be even smaller than the intra-subject variation. 
The organ intensity may also be related to the slice cross-sectional position. For instance, kidneys are not homogeneous structures, consisting of sub-structure such as renal medulla, renal papilla, renal pyramids, etc. The density of each position is different, and thus the intensity is different. For the homogeneous organs such as the liver and spleen, if the 2D slices are taken in the marginal part of the organ, the organ may have a blurry boundary which may introduce the intensity difference. 
Within the subject, the area of the abdominal wall and body mask do not vary largely indicated by Figure~\ref{fig:fig3}, which can help explain the high ICC for the area. However, the intensity of the abdominal wall and body mask is mainly depended on the amount and type of soft tissues inside, which may also be affected by slice cross-sectional position.

Herein, we propose an abdominal tissues and organ segmentation pipeline and perform a low-dose single slice CT images longitudinal variation analysis on 300 subjects. Our results show that our methods are stable on longitudinal scans and muscle, abdominal wall, fat and body mask have low longitudinal variation within two years. However, organs and other tissues that primarily related to the slice cross-sectional position may not provide sufficient evidence to perform the longitudinal variability analysis.

\section{Acknowledgments}
This research is supported by NSF CAREER 1452485 and the National Institutes of Health (NIH) under award numbers R01EB017230, R01EB006136, R01NS09529, T32EB001628, 5UL1TR002243-04, 1R01MH121620-01, and T32GM007347; by ViSE/VICTR VR3029; and by the National Center for Research Resources, Grant UL1RR024975-01, and is now at the National Center for Advancing Translational Sciences, Grant 2UL1TR000445-06. This project was also supported by the National Science Foundation under award numbers 1452485 and 2040462. This research was conducted with the support from the Intramural Research Program of the National Institute on Aging of the NIH. The content is solely the responsibility of the authors and does not necessarily represent the official views of the NIH. This study was in part using the resources of the Advanced Computing Center for Research and Education (ACCRE) at Vanderbilt University, Nashville, TN. The identified datasets used for the analysis described were obtained from the Research Derivative (RD), database of clinical and related data.

\bibliography{report} 
\bibliographystyle{spiebib} 

\end{document}